\pdfoutput=1
\documentclass[11pt]{article}
\usepackage{amsmath}
\usepackage{amsfonts}
\usepackage[final]{acl}
\usepackage{times}
\usepackage{latexsym}
\usepackage{tikz}
\usetikzlibrary{positioning, arrows.meta, calc}
\usepackage[T1]{fontenc}
\usepackage{booktabs}

\usepackage[utf8]{inputenc}

\usepackage{microtype}

\usepackage{inconsolata}

\usepackage{graphicx}

\title{SMOTExT: SMOTE meets Large Language Models}

\author{
  \textbf{Mateusz Bystroński\textsuperscript{1}},
  \textbf{Mikołaj Hołysz\textsuperscript{1}},
  \textbf{Grzegorz Piotrowski\textsuperscript{1}},\\
  \textbf{Nitesh V. Chawla\textsuperscript{2}},
  \textbf{Tomasz Kajdanowicz\textsuperscript{1}}
\\
\textsuperscript{1}Wrocław University of Science and Technology,
\textsuperscript{2}University of Notre Dame
\\
\small{
\textbf{Correspondence:} \href{mailto:mateusz.bystronski@pwr.edu.pl}{mateusz.bystronski@pwr.edu.pl}
}
}

\begin{document}
\maketitle
\begin{abstract}
Data scarcity and class imbalance are persistent challenges in training robust NLP models, especially in specialized domains or low-resource settings. We propose a novel technique, \textbf{SMOTExT}, that adapts the idea of Synthetic Minority Over-sampling (SMOTE) to textual data. Our method generates new synthetic examples by interpolating between BERT-based embeddings of two existing examples and then decoding the resulting latent point into text with xRAG architecture. By leveraging xRAG’s cross-modal retrieval-generation framework, we can effectively turn interpolated vectors into coherent text. While this is preliminary work supported by qualitative outputs only, the method shows strong potential for knowledge distillation and data augmentation in few-shot settings. Notably, our approach also shows promise for privacy-preserving machine learning: in early experiments, training models solely on generated data achieved comparable performance to models trained on the original dataset. This suggests a viable path toward safe and effective learning under data protection constraints.

\end{abstract}

\section{Introduction}

Building reliable NLP models in low-resource scenarios or for under-represented classes remains a major challenge due to the limited availability of annotated training data and ethical concerns. Data augmentation offers a promising strategy to mitigate this issue by generating additional synthetic examples. However, common textual augmentation techniques—such as synonym replacement, random deletion, or back-translation—often introduce semantic drift, grammatical errors, or distributional inconsistencies. In contrast, interpolation-based augmentation has proven effective in continuous feature spaces, especially in domains like vision or tabular data. The \textbf{S}ynthetic \textbf{M}inority \textbf{O}ver-sampling \textbf{TE}chnique (SMOTE) \cite{Chawla2002} is a well-known example that creates new minority-class samples by linearly interpolating between feature vectors. Adapting such interpolation to text, however, is non-trivial due to the discrete and high-dimensional nature of natural language.
\begin{figure}[t]
    \centering
    \includegraphics[width=0.9\linewidth]{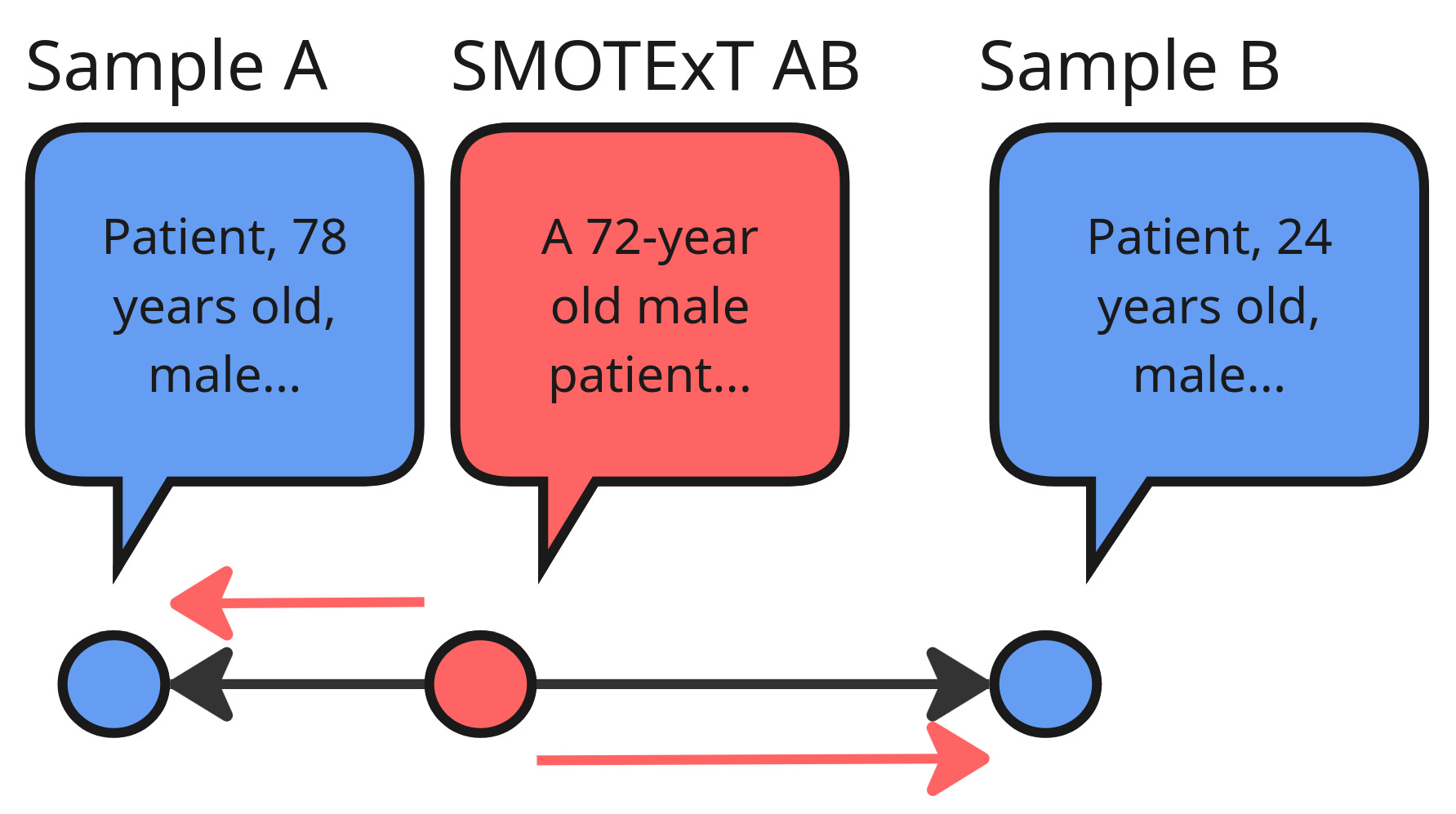}
    \caption{Two textual examples A and B are encoded into embeddings. An interpolated vector SMOTExT is fed into the xRAG model via its modality bridge (projector). Language model then generates a synthetic text that reflects combined features of the inputs.}
    \label{fig:architecture}
\end{figure}
We propose a novel method—\textbf{SMOTExT}—that performs SMOTE-style interpolation for text by operating in a continuous latent space. Specifically, we embed two textual examples using a pre-trained encoder and compute a linear interpolation between their embeddings. This intermediate vector is then decoded into natural language using xRAG \cite{Cheng2024}.

This approach enables the generation of realistic, domain-consistent samples that can enrich training data in few-shot or imbalanced-class settings. Compared to perturbation-based augmentation, interpolation in latent space may better preserve semantic relationships between examples. However, interpolated vectors are not guaranteed to remain on the natural language manifold, and out-of-distribution artifacts may occur—particularly when inputs differ significantly. Nevertheless, by decoding these vectors using a strong generative model, we can often obtain fluent and contextually appropriate text, especially when source examples are from the same domain or class.

Our contributions are as follows: (1) We introduce a novel textual data augmentation technique based on embedding space interpolation, inspired by SMOTE and enabled by large-scale pre-trained models; (2) We provide a real-world proof-of-concept in the medical domain, showing that our method can synthesize medically plausible patient descriptions; (3) We conduct a preliminary quantitative evaluation on a general-purpose classification task, showing that the method improves metrics in low-resource settings and approximates real data distribution when used alone.   and (4) We discuss limitations of the current approach and outline directions for future quantitative evaluation, including applications in privacy-preserving and generative NLP tasks.

\section{Related Work}
\paragraph{Text Data Augmentation.} Numerous augmentation methods address data scarcity in NLP. Lexical approaches like EDA \cite{Wei2019} apply random synonym replacement, insertion, or deletion, but often risk semantic drift or ungrammaticality. Paraphrasing via back-translation \cite{Sennrich2016} or prompt-based LLM generation offers more fluency, though may drift off-distribution without constraints.

Other techniques operate in continuous representation spaces. Mixup \cite{Zhang2018,Verma2019}, originally for vision, has been adapted to NLP by interpolating hidden states. TMix \cite{Chen2020} and SSMix \cite{Yoon2021} create intermediate representations by mixing hidden layers in transformer models, improving training. However, these methods are typically tied to specific architectures and do not generate standalone text.

In contrast, our method decouples augmentation from model training by decoding interpolated embeddings into human-readable text. This enables model-agnostic data augmentation and facilitates qualitative inspection of synthetic examples.

\paragraph{SMOTE and Interpolation.} Our approach is inspired by SMOTE \cite{Chawla2002}, a classic oversampling technique that mitigates class imbalance by interpolating between a minority class instance and one of its neighbors in feature space. This generates synthetic samples within the minority region, avoiding duplication. While effective in continuous domains (e.g., numeric or image data), applying SMOTE to text requires a suitable latent representation.

DeepSMOTE \cite{deepSMOTE} extended this idea to high-dimensional data by performing interpolation in the latent space of a neural encoder. It maps images into a semantic embedding space, applies SMOTE-style interpolation, and reconstructs synthetic examples via a decoder—circumventing unrealistic blending in pixel space. We adopt a similar principle for text, operating directly in the embedding space of pre-trained language models.

\paragraph{xRAG.} The recently proposed xRAG model \cite{Cheng2024} enables language models to condition on a single dense vector, derived from a retrieved document, via a special token. A fixed retriever generates document embeddings, which are projected into the LM’s input space using a small trainable \textit{modality bridge}. Only this projection is trained, keeping both the retriever and the LM frozen. The LM attends to the resulting embedding and generates fluent, context-aware output.

\section{Proposed Method}
Our approach leverages xRAG’s capability to integrate and decode an arbitrary fixed-length embedding into natural language. Essentially, we repurpose the xRAG architecture as a decoder for interpolated example embeddings. Prior work \cite{cai2021isotropy} has shown that BERT \cite{Devlin2019} embeddings tend to form well-structured and concentrated clusters in the latent space, making linear interpolation between them a viable strategy for generating semantically meaningful intermediate representations. This structural property supports the idea that interpolated points in embedding space can correspond to plausible language outputs. xRAG provides an off-the-shelf, pre-trained mechanism to realize such vectors as text, without the need to train a new autoencoder or decoder architecture from scratch. By treating the interpolated vector as if it were a retrieved document embedding, we enable the language model to generate a coherent sentence that blends features of the source examples.

Our method, consists of three main steps: (1) obtaining vector representations of input texts; (2) interpolating between two such vectors to produce a new latent point; and (3) decoding this latent vector into text using xRAG. 

%\begin{figure}[t]
%    \centering
%    \rule{0.9\linewidth}{0.5\linewidth} % Placeholder box for the architecture diagram
%    \caption{Overview of the proposed approach. Two textual examples $x_i$ and $x_j$ are encoded into embeddings $e_i$ and $e_j$. An interpolated vector $e_{\text{new}} = \lambda e_i + (1-\lambda) e_j$ is fed into the xRAG model via its modality bridge (projector). xRAG's frozen language model then generates a synthetic text $y_{\text{new}}$ that reflects combined features of the inputs.}
%    \label{fig:architecture}
%\end{figure}

\paragraph{Embedding Text Examples.} We first convert each text example into a fixed-dimensional embedding. In this work we use the [CLS] representation from \texttt{SRF-Embedding-Mistral} \cite{SFRAIResearch2024}. Let $x_i$ and $x_j$ be two text examples (e.g., two sentences or short paragraphs). We obtain $e_i = \text{BERT-Enc}(x_i)$ and $e_j = \text{BERT-Enc}(x_j)$, where $e_i \in \mathbb{R}^d$ (with $d=4096$ in our case). These embeddings are intended to capture the major features of each text in a continuous vector form.

\paragraph{Latent Space Interpolation.} Next, we generate a new latent point between $e_i$ and $e_j$. We sample a mixing coefficient $\lambda \in [0,1]$. We then compute the interpolated vector $e_{\text{new}} = \lambda e_i + (1-\lambda) e_j$. This linear interpolation in the embedding space is analogous to the DeepSMOTE procedure in image data, which assumes that the space is locally linear such that points between two same-class examples are likely to remain within the same class region. In practice, $e_{\text{new}}$ should represent a blend of the semantic content from $x_i$ and $x_j$. One key assumption here is that the embedding space of the encoder is smooth enough that this linear combination yields a meaningful intermediate representation, rather than an out-of-distribution vector. We also note that to maintain label consistency (in a supervised setting), $x_i$ and $x_j$ would typically be chosen from the same class or category.

\paragraph{Decoding via xRAG.} The core novelty of our approach lies in using a xRAG architecture to decode $e_{\text{new}}$ into a synthetic text $y_{\text{new}}$. In our work we use xRAG-7B model \cite{Cheng2024} that was trained to accept an embedded context vector through a special token and generate an output conditioned on it. We denote the language model component as $\text{LM}_{\text{xRAG}}$ and the modality bridge (projector) as $W_p$. $W_p$ is a linear projection that takes as input a context embedding (originally, a document embedding from the retriever) and outputs a vector in the same space as the LM’s token embeddings. During xRAG’s training, $W_p(e)$ is inserted into the LM’s input sequence as the embedding for a designated token (let's call it [X]). This way, the LM can attend to [X] as if it were a compressed representation of a document relevant to the task (e.g., answering a question or providing knowledge).

To use xRAG for our purpose, we feed the interpolated embedding through the same projector and prompt the LM to paraphrase the decoded text. Specifically, we compute $h_X = W_p(e_{\text{new}})$. Because $\text{LM}_{\text{xRAG}}$ was trained with an autoencoding objective on documents—particularly in the context of paraphrasing—it has learned to generate coherent text that reflects the semantic content of $h_X$..

\paragraph{Example Workflow.} Pseudocode for generating one synthetic example is as follows:
\begin{quote}\small
\texttt{Input: $x_i$, $x_j$ (text examples)}\\
\texttt{$e_i \leftarrow \text{BERT-Enc}(x_i)$}\\
\texttt{$e_j \leftarrow \text{BERT-Enc}(x_j)$}\\
\texttt{Sample $\lambda \in [0,1]$}\\
\texttt{$e_{\text{new}} \leftarrow \lambda e_i + (1-\lambda) e_j$}\\
\texttt{$h_X \leftarrow W_p(e_{\text{new}})$ \quad // project to LM space}\\
\texttt{Feed $h_X$ as [X] into $\text{LM}_{\text{xRAG}}$, generate $y_{\text{new}}$}
\end{quote}

In practice, multiple synthetic samples can be generated by choosing different pairs $(x_i, x_j)$ and/or different $\lambda$ values. These $y_{\text{new}}$ outputs can then be included as additional training data for downstream models or further analyzed.
\section{Experiments}
\begin{table*}
  \centering
  \begin{tabular}{lccc}
    \hline
    \textbf{Training Setup} & \textbf{Accuracy} & \textbf{Macro F1} & \textbf{Weighted F1} \\
    \hline
    Real only (1,000 samples)        & 0.8361 & 0.8382 & 0.8382 \\
    Real + SMOTExT (1,000 + 2,600) & \textbf{0.8394} & \textbf{0.8404} & \textbf{0.8408} \\
    SMOTExT only (2,600 samples)   & 0.8037 & 0.8038 & 0.8036 \\
    \hline
  \end{tabular}
  \caption{\label{tab:results}
    Classification results on a low-resource 20 Newsgroups subset. Augmenting real training data with 2,600 interpolated SMOTExT examples improves macro and weighted F1 scores. Training on synthetic data alone approximates the performance of the real-data baseline, suggesting potential for privacy-preserving dataset generation.
  }
\end{table*}
To provide a proof of concept for the proposed interpolation-based text augmentation method, we conduct a set of small-scale experiments designed to test its viability in low-resource classification settings. While these evaluations are preliminary, they offer insights into the effectiveness and limitations of the approach and help establish a foundation for future, more comprehensive benchmarking.

We include in Appendix~\ref{sec:appendix-clinical} a clinical case study illustrating how our method generates semantically coherent synthetic samples by interpolating between real examples in the latent space. The resulting text combines features from both inputs while maintaining domain plausibility.

To further evaluate the proposed method in a general-purpose classification setting, we use the \texttt{20 Newsgroups} dataset.\footnote{\url{https://scikit-learn.org/0.19/datasets/twenty_newsgroups.html}} This corpus contains discussion posts from 20 topical newsgroups that form several well-known semantic clusters. Following \texttt{scikit-learn}'s recommendations, we remove email headers, signatures, and quoted content to reduce information leakage and simplify the classification task.

We select a subset of classes that are semantically distant from each other, as prior work has shown that some groups (e.g., computer-related or religion-related) are challenging to distinguish—even for human annotators. As a baseline, we fine-tune a \texttt{ModerBERT} \cite{warner2024smarterbetterfasterlonger} model, freezing all encoder layers and training only the classification head. 

To simulate a low-resource scenario, we restrict the training set to 1,000 real examples. We then generate 2,600 SMOTExT examples using our latent interpolation pipeline and add them to the training data. We compare three conditions: (1) training on 1,000 real examples (baseline), (2) training on 1,000 real + 2,600 SMOTExT examples, and (3) training on 2,600 SMOTExT examples alone. Results are reported in Table~\ref{tab:results}.

The addition of SMOTExT samples yields modest but consistent improvements in both macro and weighted F1 scores. This demonstrates the usefulness of interpolation-based augmentation in low-resource settings.

Notably, training on SMOTExT data alone results in competitive performance, with only a small drop relative to the real-data baseline. This suggests that the SMOTExT samples capture meaningful features of the original distribution—an encouraging result for privacy-sensitive scenarios. The slight performance gap also highlights a known limitation of generative approaches based on latent interpolation: without constraints, some outputs may drift slightly out of distribution, introducing minor inconsistencies or irrelevant details. Addressing this challenge will be an important focus of future work.

\section{Discussion and Conclusions}
We presented a method to synthesize new textual data points by interpolating between sentence embeddings and decoding the result using a large language model. This approach adapts the core idea of SMOTE to the NLP setting.

Our preliminary results show that the method can generate fluent, coherent texts that plausibly combine attributes of their source inputs. This suggests potential utility in data augmentation for low-resource or imbalanced classification tasks, as well as in generative scenarios where realism and contextual richness matter more than label precision. Importantly, since the method operates purely on latent representations and decodes them into novel outputs, it offers a promising direction for privacy-preserving data synthesis—particularly in sensitive domains such as healthcare, where access to real patient records is often restricted.

Beyond augmentation, our approach also opens up the possibility of generating fully synthetic datasets that approximate the original data distribution. This is particularly promising for training generative models, where access to large, high-quality datasets is essential, yet often constrained by privacy, legal, or ethical considerations. By distilling the structure and diversity of real data into novel synthetic samples, our method may enable pretraining or fine-tuning of generative language models without direct exposure to protected content. This has the potential to mitigate risks of data leakage while preserving downstream performance.

Importantly, our method does not rely on task-specific training: synthetic data can be generated once and reused across applications. This makes it particularly attractive for semi-structured tasks such as generating medical histories, populating patient profiles, or augmenting dialogue datasets. We believe this approach has wide applicability across domains and task types, especially where labeled data is scarce.

Future work will focus on scaling the evaluation to more domains and tasks, exploring interpolation beyond pairs (e.g., multi-point or manifold-based mixing), and developing mechanisms for better control over semantic consistency and out-of-distribution risks. We also plan to investigate strategies for automatic filtering or calibration of generated examples.

\section{Limitations}
\label{sec:limitations}

This work is in an early exploratory stage, but initial experiments provide encouraging evidence. Nonetheless, several limitations remain.

First, while we now present preliminary quantitative results, a more comprehensive evaluation across a wider range of tasks and domains is needed to fully assess the generality and impact of our approach. Our current benchmarks are limited to a subset of the 20 Newsgroups dataset, and further analysis is required to determine how well the method performs in truly low-resource or high-stakes settings such as medicine or law.

Secondly, although we assume that interpolating between two examples from the same class yields a valid sample from that class, this assumption is not formally guaranteed and may result in out-of-distribution samples. Nevertheless, our experiments demonstrate that a model trained exclusively on SMOTExT data achieves performance comparable to that of a model trained on real examples, suggesting that SMOTExT effectively approximates the underlying data distribution.

A further limitation of our current setup is the use of a relatively small language model (7B parameters). While this choice enables efficient experimentation, it may limit the expressiveness and fidelity of the generated outputs. Nevertheless, the method proves effective even in this setting, and we expect that applying it with larger models would lead to improved results.

Lastly, while the method shows promise for privacy-preserving synthetic data generation—particularly by avoiding direct exposure to sensitive inputs—it currently lacks explicit mechanisms to enforce semantic constraints or prevent subtle inconsistencies. In generative applications, this may be acceptable; but in settings where factual accuracy or clinical safety is paramount, stricter control mechanisms will be necessary.

\bibliography{custom}

\appendix
\section{Appendix}
\label{sec:appendix}

\subsection{Clinical Text Interpolation Case Study}
\label{sec:appendix-clinical}

To illustrate the generative capabilities of our method in a structured domain, we present a proof-of-concept example in the medical setting. Due to privacy concerns and annotation costs, medical text datasets are typically small, making them well-suited for data augmentation.

We constructed two clinical cases (A and B) using GPT-4o to simulate realistic outpatient scenarios. These were embedded using a BERT-based encoder, linearly interpolated in latent space, and decoded using xRAG-7B to produce a SMOTExT case A-B.

\begin{table}[h]
\centering\small
\begin{tabular}{p{0.95\linewidth}}
\toprule
\textbf{Example A (Original)}:\\ 
\emph{``Patient, 24 years old, male. No significant chronic illnesses. Presents with mild sore throat for the past 2 days. No fever, no shortness of breath. No known allergies. Non-smoker, not on any regular medications.''}\\[1ex]
\textbf{Example B (Original)}:\\ 
\emph{``Patient, 78 years old, male. Type 2 diabetes, hypertension, history of myocardial infarction 5 years ago. Complains of exertional dyspnea and dry cough for the past week. Temperature 37.8\textdegree C. Takes metformin, bisoprolol, and ramipril.''}\\[1ex]
\textbf{SMOTExT A-B (Synthetic)}:\\ 
\emph{``A 72-year-old male patient presented to the clinic with a chief complaint of shortness of breath and chest pain for the past 2 days. He described the symptoms as a feeling of tightness in the chest, which worsened with exertion and improved with rest. The patient also reported a non-productive cough and mild fever. He denied any recent travel, contact with sick individuals, or exposure to allergens. The patient's past medical history was significant for hypertension, hyperlipidemia, and type 2 diabetes mellitus. He was a former smoker, having quit 10 years ago. The patient's vital signs were within normal limits, and his physical examination was unremarkable. His laboratory results showed an elevated white blood cell count, normal electrolytes, and a slightly elevated creatinine level. A chest X-ray was ordered, which revealed a possible infiltrate in the right lower lobe. The patient was diagnosed with a possible respiratory infection and was prescribed antibiotics and a bronchodilator. He was advised to follow up with his primary care physician for further evaluation and management.''}\\
\bottomrule
\end{tabular}
\caption{Proof-of-concept generation via latent interpolation in the medical domain. The SMOTExT case plausibly integrates features from both source cases (A and B) while maintaining clinical coherence.}
\label{tab:appendix-example}
\end{table}

\begin{table*}
  \centering
  \begin{tabular}{p{0.20\linewidth}p{0.25\linewidth}p{0.25\linewidth}p{0.25\linewidth}}
    \hline
    \textbf{Attribute} & \textbf{Example A} & \textbf{Example B} & \textbf{SMOTExT A-B} \\
    \hline
    Age & 24 years old & 78 years old & 72 years old (closer to B) \\
    Chronic conditions & None & Type 2 diabetes, hypertension, myocardial infarction & Hypertension, type 2 diabetes, hyperlipidemia (mix of B + novel) \\
    Symptom profile & Mild sore throat, no fever or dyspnea & Exertional dyspnea, dry cough, fever (37.8°C) & Shortness of breath, chest pain, dry cough, mild fever (onset similar to A, symptoms closer to B) \\
    Medications & None & Metformin, bisoprolol, ramipril & Not specified \\
    Lifestyle factors & Non-smoker & Not mentioned & Ex-smoker (new plausible feature) \\
    \hline
  \end{tabular}
  \caption{\label{tab:comparison}
    Structured comparison of two original clinical descriptions (A and B) and a SMOTExT casE generated via latent space interpolation. The SMOTExT case plausibly combines features from both inputs while maintaining medical realism and narrative fluency.
  }
\end{table*}

This comparison highlights how the generated sample semantically resides between the two source cases, blending demographic, symptomatic, and comorbidity information. The resulting case maintains medical plausibility and exhibits coherent narrative structure. 

The example was reviewed by a physician, who confirmed its consistency with medical knowledge. Minor suggestions included the possible addition of elevated temperature and auscultatory findings in the lungs to fully match the described clinical picture. Nonetheless, the overall scenario was deemed realistic and clinically valid.

\end{document}